\documentclass{article}
\usepackage{graphicx} %
\usepackage{subcaption} 
\usepackage{caption} 

\usepackage[preprint,nonatbib]{neurips_2024}

\usepackage{amsmath}
\usepackage{amsthm}

\theoremstyle{definition}

\usepackage[utf8]{inputenc} 
\usepackage[T1]{fontenc}    
\usepackage{hyperref}       
\usepackage{url}            
\usepackage{booktabs}       
\usepackage{amsfonts}       
\usepackage{nicefrac}       
\usepackage{microtype}      
\usepackage{xcolor}         
\usepackage{comment}

\title{High precision PINNs in unbounded domains: application to singularity formulation in PDEs}

\author{%
Yixuan Wang$^{1}$\thanks{roywang@caltech.edu} \quad Ziming Liu$^{2,3}$ \quad Zongyi Li$^{1}$ \quad Anima Anandkumar$^{1}$\quad Thomas Y. Hou$^{1}$ \\
$^1$ California Institute of Technology \\ $^2$ Massachusetts Institute of Technology \\ $^3$ The NSF Institute for Artificial Intelligence and Fundamental Interactions
}

\begin{document}

\maketitle

\begin{abstract}
We investigate the high-precision training of Physics-Informed Neural Networks (PINNs) in unbounded domains, 
with a special focus on applications to singularity formulation in PDEs. 
We propose a modularized approach and study the choices of neural network ansatz, sampling strategy, and optimization algorithm.
When combined with rigorous computer-assisted proofs and PDE analysis,
the numerical solutions identified by PINNs, provided they are of high precision, 
can serve as a powerful tool for studying singularities in PDEs.
For 1D Burgers equation, our framework can lead to a solution with very high precision,
 and for the 2D Boussinesq equation, which is directly related 
 to the singularity formulation in 3D Euler and Navier-Stokes equations, 
 we obtain a solution whose loss is $4$ digits smaller than that obtained in \cite{wang2023asymptotic} with fewer training steps. 
  We also discuss potential directions for pushing towards machine precision for higher-dimensional problems.

\end{abstract}
 
\section{Introduction}
Singularity formulation is one of the key challenges in the study of partial differential equations (PDEs).
Unlike well-posed equations, where one can apply classical existence and uniqueness theorems,
singularities often occur in certain solutions of nonlinear PDEs, where we only have guarantees 
of existence for a short time, but the solution may blow up in finite time.
The study of singularities often involves a case-by-case approach and is related to 
some of the most intriguing mathematics and physical properties, such as the onset of turbulence in the Navier-Stokes equations.
The singularity of the Navier-Stokes equations is one of the seven Millennium Prize Problems \cite{fefferman2006existence}:
widely regarded as the most fundamental and challenging problem in analysis, and is still open.
One of the key difficulties in the study of singularities is the lack of 
understanding of the singularity pattern and its mechanism. The computation of the 
singularity itself, or infinity, is intractable numerically.
A general roadmap is thus to first propose a plausible singularity ansatz that renders
the computation feasible,
then find candidates of such blowup by numerical simulations, and
finally verify the stability of such an ansatz by PDE analysis.

We are often interested in a special structure of singularity: self-similar singularity.
Self-similarity relates to the invariance of the solution under scaling transformations 
and reduces singularity to the existence of a self-similar profile. To be precise, for the quantity of interest $u(x,t)$,
we put the ansatz $u(x,t) = (T-t)^{-\alpha} U(x(T-t)^{-\beta})$, where $U$ is the profile function
independent of time, $T$ is the blowup time, and $\alpha>0, \beta$ are the scaling exponents to be determined.
Now we can reduce the computation of an infinite $u$ to the computation of a finite, smooth profile $U$, 
along with scaling exponents $\alpha, \beta$ to be inferred. Physics-Informed Neural Networks (PINNs) \cite{raissi2019physics}
serve as a powerful tool to find such profiles, with the scaling parameters jointly inferred as inverse problems. It was first introduced in the context of identifying singularity profiles in \cite{wang2023asymptotic} and has seen success 
even in problems that are unstable for traditional numerical methods. While PINNs offer a powerful tool to search
for candidate profiles, solutions identified by PINNs are often of limited accuracy and far from applicable to
rigorous PDE analysis, to the best of the authors' knowledge.

In this work, we aim to systematically study the high-precision training of PINNs,
with a special focus on applications  to solve profile equations governing singularity formulations
in PDEs, in fluid dynamics and beyond. We will take a modularized perspective without diving 
into sophisticated tricks and investigate the following aspects:
a good neural network ansatz representing the profile function, a good sampling strategy to tackle
the infinite domain with a special focus on imposing boundary conditions,
and a good optimization algorithm to train the neural network. 
We apply our findings to the 1D Burgers equation and the 2D Boussinesq equation, 
obtaining a precision amenable to rigorous PDE analysis for the 1D Burgers equation 
and $4$ digits better than \cite{wang2023asymptotic} with fewer training steps for the 2D Boussinesq equation.
The 2D Boussinesq equation shares many similarities with the 3D axisymmetric Euler equation for the ideal
fluid without viscosity; see the pioneering works of \cite{chen2021finite,chen2022stable,chen2025stable} for the connection between the two equations,
where the authors used the connection to establish singularity formulation for the 3D axisymmetric Euler with boundary.

\section{Related Works}
\subsection{PINNs}
Neural networks have witnessed success in solving PDEs and surrogate modeling in math and science.
PINNs in particular have been widely used due to their flexibility and applicability to a wide range of problems \cite{cai2021physics, raissi2019physics, karniadakis2021physics}.
The key idea of PINNs is to enforce the PDE constraints at a set of collocation points and to minimize the residual
of the PDEs as a loss function. By posing the solution of PDEs as an optimization problem, PINNs are especially suited
to solve inverse problems \cite{raissi2020hidden, yuan2022pinn, lu2021physics, yu2022gradient}, where the solution of the PDE and the underlying parameters can be jointly inferred.
In \cite{wang2023asymptotic}, the authors used PINNs to study the blowup of the 1D Burgers equation, 
the 1D family of generalized Constantin-Lax Majda equations, and the 2D Boussinesq equation.

Another line of work, operator learning \cite{kovachki2023neural}, focuses on learning the solution operator instead of learning 
a single instance of solution, where Fourier Neural Operators (FNOs) \cite{li2020fourier, li2023fourier, li2023geometry, li2024scale} and DeepONets \cite{lu2021learning} are two families of representative works in this direction.
Once a solution operator is learned, it can be evaluated in a resolution-free manner at any point in the domain.
Data are often augmented to enhance the solution accuracy, while the loss function  can also incorporate
the PDE constraints, termed Physics Informed Neural Operator (PINO) \cite{li2021physics}. In \cite{maust2022fourier}, the authors used PINO with Fourier continuation 
to study the blowup of the 1D Burgers equation.
\subsection{Self-similar Singularity and Computer Assisted Proofs}
Self-similar singularity of the ansatz $u(x,t) = (T-t)^{-\alpha} U(x(T-t)^{-\beta})$
is generic in the study of singularity formation in PDEs, where 
one uses the scaling invariance of the PDE and
 can reduce the computation of an infinite $u$ to the computation of a finite, smooth (approximate) profile $U$.
Such structures exist even in the simple Riccati ODE $u_t=u^2$ with an exact solution $u=(T-t)^{-1}$.

The approximate profile can be identified via explicit construction or numerical computation. 
Working in the rescaled, self-similar variables and performing 
stability analysis around the profile $U$ provides a powerful tool to establish the singularity formation,
for nonlinear Schrodinger equations \cite{mclaughlin1986focusing, merle2005blow}, incompressible fluids \cite{elgindi2019finite,chen2019finite2,chen2021finite,chen2022stable,hou2024blowup}, compressible fluids \cite{MRRSam22a, MRRSam22b}, and beyond. 
Until recently, most of the works relied on an explicit profile and spectral information 
of the associated linearized operator to establish linear and nonlinear stability. In
\cite{chen2022stable,chen2025stable}, the authors used computer-assisted proofs with a sophisticated numerical profile obtained by solving the dynamic rescaling equations in time to obtain an approximate steady state. By analyzing the stability of the approximate profile, they established the singularity formation for the 2D Boussinesq equation
and the 3D axisymmetric Euler equation with boundary.
And in \cite{hou20242,chen2024stability}, the authors provided a framework using only local information for stability analysis,
bypassing spectral information and allowing for numerical profiles with computer-assisted proofs,
for problems beyond self-similarity.

\subsection{Towards High Precision Training}
Various methods have been proposed in the literature to improve the accuracy of PINNs. One line of work focuses on a 
better representation of the solution. In \cite{michaud2023the, wang2024multi}, the boosting technique was proposed, where a sum of a sequence of neural networks
with decreasing magnitude was used to learn the solution; at each stage, a new neural network is trained to learn the residual.
To overcome the spectral bias \cite{rahaman2019spectral} of multilayer perceptrons (MLPs), or the favor of learning low-frequency modes \cite{xu2019training,xu2019frequency} in the solution, 
one can use Fourier feature encoding \cite{sitzmann2020implicit, tancik2020fourier,ng2024spectrum}, or different activation functions \cite{jagtap2020adaptive, jagtap2020locally,hong2022activation, zhang2023shallow,wang2024expressiveness}. In particular, Kolmogorov-Arnold Networks (KANs) \cite{liu2024kan,liu2024kan2} that leverage nonlinear learnable activation functions and the Kolmogorov-Arnold representation theorem were proposed and further investigated in the PINN setting \cite{wang2024kolmogorov,shukla2024comprehensive,toscano2025pinns}.
Another line of work improves the optimization landscape during the training of PINNs.
Various optimizers, which we will detail in Subsection \ref{subsec:optimizer}, have been proposed to improve the convergence rate.
Adaptive design of points sampling \cite{anagnostopoulos2024residual,wu2023comprehensive,rigas2024adaptive} and adaptive weighting of different terms \cite{wang2021understanding,xiang2022self,mcclenny2023self} in the loss function
were also proposed to improve the accuracy of PINNs.

We will only focus on applying hard constraints and choosing a good optimizer in this work and leave 
the exploration of more sophisticated tricks for future work.
\section{Methodology} 
We outline our methodology of high-precision training for PINNs on the whole space in this section.
We work under the general formulation of the profile equation $$L(U,\lambda)=0,$$ where 
$U(y)$ is the profile function, $\lambda$ is a set of scaling parameters to be determined,
and $L$ is the nonlinear differential operator. For our problems of interest, $U$=0 will be a trivial 
solution satisfying the equation.
\subsection{Infinite Domain}\label{subsec:inf}
The key challenges we are facing here are sampling and learning on an infinite domain.
For a given budget of a finite number of sampling points, we need to sample the domain in a way that
the resulting solution is accurate and generalizes well throughout the domain. In the meantime,
We want the neural network to be able to represent the profile function and the initialization of parameters
to favor learning of such representations. To this end, we adopt an exponential "mesh" in our sampling strategy:
Consider an auxiliary  variable $z$ such that $y = \sinh(z)=\frac{e^z-e^{-z}}{2}$, 
and sample $z$ uniformly in a finite region. Here we choose the $\sinh$ transformation as in \cite{wang2023asymptotic}
to respect the parity of the functions, detailed in the subsequent subsection. Such a transformation
maps roughly $z\in[-30,30]$ to $y\in[-5\times 10^{12},5\times 10^{12}]$.

Boundary conditions are another important aspect when learning on the whole space.
For our problems of interest, $U$ by itself will not have sufficient decay at infinity,
and one approach adopted in \cite{wang2023asymptotic} is to impose Neumann boundary conditions at infinity, or numerically 
on the boundary of the domain of the $z$ variables.
To rule out the trivial solution $U=0$, we need to enforce a nondegeneracy condition,
often posed at the origin. We will discuss the enforcement of these conditions in the following subsection.
We refer to this formulation as boundary conditions using \textbf{weak asymptotics}.

On the other hand, we can enforce stronger information on the boundary. 
If we know the exact asymptotic behavior of the solution at infinity as $g$, for example a power law,
we can introduce a smooth cutoff function $\chi$ with $\chi(0)=0,\chi(\infty)=1$
 and the ansatz $U=\tilde{U}+\chi g$. We can then enforce Dirchlet boundary conditions at infinity for $\tilde{U}$
 represented by the neural network. We refer to this formulation as boundary conditions using \textbf{exact asymptotics}. We will demonstrate for the 1D example that PINNs using exact asymptotics will outperform those using weak asymptotics by a large margin.

A priori, the exact asymptotics information is not available, and one can first train a neural network $U_w$ with 
boundary conditions using weak information, and distill the information of asymptotics $g$ from $U_w$.
We refer to this formulation as boundary conditions using \textbf{hybrid asymptotics}. For example,
for the 2D Boussinesq equation, borrowing ideas from \cite{chen2022stable},
one can use function fitting and symbolic regression to extract asymptotics $g$ from $U_w$, filtering out the noisy residues, such that $g$ is a symbolic function approximating $U_w$ at infinity. We will leave this approach to future work.

\subsection{Hard Constraint}\label{subsec:hard}
Hard constraints are important concepts in the parametrization of the solution space for PINNs.
When enforced properly, they will guarantee physical properties of the solution \cite{lu2021physics, richter2022neural,mohan2023embedding,duruisseaux2024towards}, and
can impose the solution to be in the correct manifold. While most of the previous works focus on hard constraints of boundary conditions,
we emphasize the enforcement of hard constraints in the following senses: the parity of the learned function
and the nondegeneracy conditions. Empirically we observe a better convergence rate and a more stable solution
when enforcing hard constraints.

\paragraph{Parity.} For a function $f(y_i,\hat{y}_i)$ even/odd in the variable $y_i$, we train a neural network 
with the following ansatz $f=(f_{nn}(y_i,\hat{y}_i)\pm f_{nn}(-y_i,\hat{y}_i))/2$. 

\paragraph{Nondegeneracy conditions.} As discussed in the previous subsection, we need to enforce 
nondegeneracy conditions to rule out the trivial solution $U=0$ when using weak asymptotics.
For example, for the 1D Burgers equation, we know that $U$ is odd and necessarily $U'(0)=-1$; 
we can enforce $U'''(0)=6$. We will enforce a hard constraint
via Taylor expansion at the origin as $U=-z+z^3+z^4U_1$, for an odd function $U_1$. 
Similarly for the 2D Boussinesq equation, we enforce $\partial_{1}\Omega(0,0)=-1$ and $\Omega$
is odd in $z_1$ via a Taylor expansion as $\Omega=-z_1+z_1z_2\Omega_1+z_1^2\Omega_2$, 
where $\Omega_1,\Omega_2$ are even and odd functions in $z_1$ respectively.

\subsection{Optimizer: Self-Scaled BFGS Methods}\label{subsec:optimizer}
A common practice of training PINNs is to use the Adam optimizer.
As a stochastic first-order method, Adam is known to be robust and efficient in training deep neural networks
and can empirically escape local minima. To further improve convergence to the minimizer, one 
can apply second-order methods with a higher convergence rate, like L-BFGS, after training with Adam for 
a few epochs. While this seems to be a gold standard in the training of PINNs \cite{rathore2024challenges}, various optimizers have been investigated,
 including variants of 
second-order quasi-Newton methods \cite{rathore2024challenges,wang2025gradient}, and optimizers using natural gradients \cite{muller2023achieving,jnini2024gauss,chen2024teng}.
We highlight and use the self-scaled BFGS methods proposed in \cite{al1998numerical,al2014broyden} and introduced to the PINNs context in \cite{urban2025unveiling,kiyani2025optimizer}.
BFGS methods use an approximation of the inverse of the Hessian matrix to precondition the gradient for the update direction. To be precise, consider the parameters $\Theta_k$ and learning rate $\alpha_k$ at step $k$, with loss function $\mathcal{J}(\Theta)$, then the update rule for $\Theta$ is $$\Theta_{k+1}=\Theta_k-\alpha_k H_k\nabla\mathcal{J}(\Theta_k).$$ Different choices of updating the approximate inverse Hessian $H_k$ lead to different optimizers, and L-BFGS in particular is a memory-efficient way for the updates by storing only vectors instead of the whole matrix.
The self-scaled BFGS methods use a scaling compared to the standard BFGS update of the inverse Hessian. More precisely, for the auxiliary variables $$\begin{aligned}
    s_k&=\Theta_{k+1}-\Theta_k,\quad y_k=\mathcal{J}(\Theta_{k+1})-\mathcal{J}(\Theta_k),\\{v}_k&=\sqrt{{y}_k \cdot H_k {y}_k}\left[\frac{{s}_k}{{y}_k \cdot {s}_k}-\frac{H_k {y}_k}{{y}_k \cdot H_k {y}_k}\right],\end{aligned}$$
    we have  for the scalers $\tau_k$ and $\phi_k$: $$H_{k+1}=\frac{1}{\tau_k}\left[H_k-\frac{H_k {y}_k \otimes H_k {y}_k}{{y}_k \cdot H_k {y}_k}+\phi_k {v}_k \otimes {v}_k\right]+\frac{{s}_k \otimes {s}_k}{{y}_k \cdot {s}_k},$$
   where the original BFGS corresponds to the choices $\tau_k=\phi_k=1$.
While this is only a simple modification of the original BFGS, the authors in \cite{urban2025unveiling} demonstrated a much improved convergence rate
across a variety of benchmarks, including the  Helmholtz equation, the nonlinear Poisson equation, the nonlinear Schrödinger equation, the Korteweg-De Vries equation, the viscous Burgers equation, the Allen-Cahn equation, 3D Navier-Stokes: Beltrami flow, and the lid-driven cavity.
We use the self-scaled Broyden methods proposed in \cite{urban2025unveiling}; see equations (13)-(23) therein for details on the choices of  $\tau_k$ and $\phi_k$.

\paragraph{On the role of minibatch training or random resampling.}
    One of the common practices when training PINNs is to use random resampling of the collocation points. This can enhance the performance of SGD-based methods like Adam empirically. However, full-batch second-order methods like BFGS with supposedly higher-order accuracy do not adapt well to random resampling since they rely on past trajectories for Hessian updates. 
 One empirical observation, as proposed in \cite{wang2024multi}, is that when one uses an optimizer with fixed resolution like BFGS, it will be able to generalize in the regions where sampling points are sufficient.
 However, in the undersampled regions, the learned solution generalizes poorly. In an abstract form, there exists a critical batchsize $N_c$, such that when $N>N_c$, fixed sampling will be preferred, while for  $N<N_c$, fixed sampling will have very bad generalization. 
 $N_c$ would depend on both the equation and the scale of the neural network.
     Empirically, we observe that roughly 10k points are sufficient for generalization with fixed training points. We resample every 1000 epochs to further reduce overfitting. See details on the choices of the batch size in the experiments section.
    \section{Experiments}
    In this section, we describe our numerical experiments on the blowup profiles for 1D Burgers equation and 2D Boussinesq Equation. The codes are available at \url{https://github.com/RoyWangyx/High-precision-PINNs-unbounded-domains-/tree/main}. When training both equations, we denote the PDE by $L(U(y))=0$ and the boundary condition by $B(U)=0$. We use auxiliary variables $z=\sinh^{-1}y$ as in Subsection \ref{subsec:inf} and consider the following combination of interior, boundary, and smoothness losses as in \cite{wang2023asymptotic}
    \begin{equation}\label{eqn_loss}
 \begin{aligned}loss&=
        0.1(L_i+L_s)+L_b\\&= 0.1(\frac{1}{N_i}\sum_{j=1}^{N_i} [\hat{L}(U_{nn}(z_j))]^2+\frac{1}{N_s}\sum_{j=1}^{N_s} |\nabla_{z_j} \hat{L}(U_{nn}(z_j))|^2)+\frac{1}{N_b}\sum_{j=1}^{N_b} [\hat{B}(U_{nn})]^2,
    \end{aligned}       
    \end{equation}
    where $U_{nn}(z)$ is supposed to approximate $\hat{U}(z)=U(y)$ in the $z$-variables, and $\hat{L}$, $\hat{B}$ denotes the PDE and the boundary condition transformed in the $z$-variables; see \cite{wang2023asymptotic} for a concrete formula for the 2D Boussinesq equation.
\subsection{Burgers Equation}
For the 1D Burgers equation \begin{equation}
    u_t+uu_x=0,
\end{equation}
consider the self-similar ansatz that respects the scaling symmetry 
\begin{equation}
    u(x,t)=(1-t)^\lambda U(y),\quad y={x}{(1-t)^{-1-\lambda}}.
\end{equation}
The profile equation for $U$ used for the PDE loss in \eqref{eqn_loss} is
\begin{equation}\label{burgers_eqn}
    -\lambda U+((1+\lambda)y+U)U_y=0.
\end{equation}
We impose an odd symmetry on $U$, and the profile equation has implicit solutions  \begin{equation}
    y+ U+CU^{1+1/\lambda}=0.
\end{equation}
for any constant $C$, as in the setting of \cite{wang2023asymptotic}, where we know that the most stable solutions correspond to $\lambda =0.5$ and there are nonsmooth solutions at e.g. $\lambda =0.4$.

In this example, we assume that we first train the neural network on a bounded domain and infer the correct $\lambda$ already, for example via the method in \cite{wang2023asymptotic}. Now we focus on fixing $\lambda$ and learn $U$ on the unbounded domain. Using an MLP with activation function $\tanh$, $4$ layers and $20$ neurons per layer and a hard constraint on parity, we use the optimizer SSBroyden1 as in \cite{urban2025unveiling} with $20000$ epochs and resampling every $1000$ epochs. $z$ is sampled uniformly on $[0,30]$ with a batchsize $10000$ for both the interior and smoothness losses, corresponding to a domain $[0,5\times 10^{12}]$ in the $y$ variables.

For the formulation using weak asymptotics as in Subsection \ref{subsec:inf}, we use the Neumann boundary condition $U_y=0$ and enforce hard constraint of nondegeneracy conditions as in Subsection \ref{subsec:hard}. For the formulation using exact asymptotics as in Subsection \ref{subsec:inf}, we use Dirichlet boundary condition $\tilde{U}=0$ and the cutoff function $\chi=(\frac{y}{1+y})^{15}$, since the far field is captured by the exact asymptotics $g=-y^{\frac{\lambda}{1+\lambda}}$.

\begin{figure}[htbp]
    \centering
    \begin{minipage}[b]{0.48\textwidth}
        \centering
        \includegraphics[width=\textwidth]{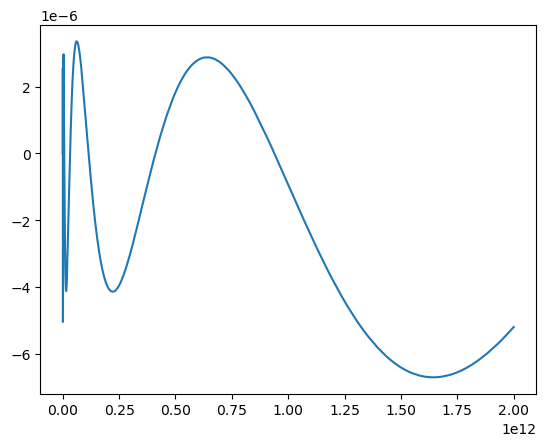}
    
    \end{minipage}
    \hfill
    \begin{minipage}[b]{0.48\textwidth}
        \centering
        \includegraphics[width=\textwidth]{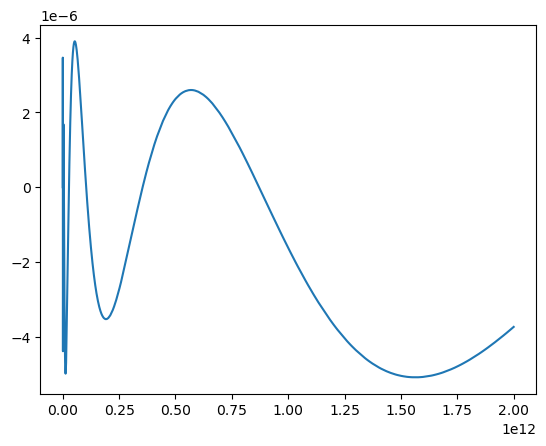}
    \end{minipage}

    \vspace{1em}

    \begin{minipage}[b]{0.48\textwidth}
        \centering
        \includegraphics[width=\textwidth]{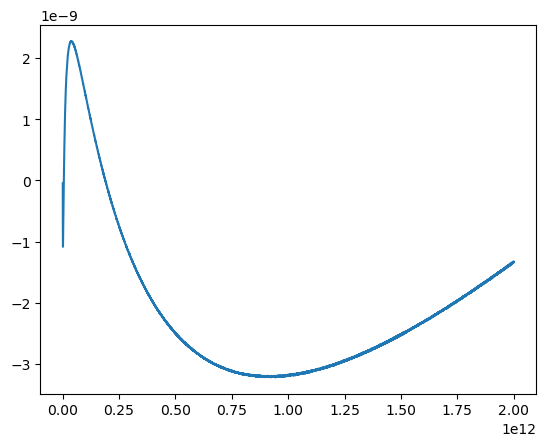}
    \end{minipage}
    \hfill
    \begin{minipage}[b]{0.48\textwidth}
        \centering
        \includegraphics[width=\textwidth]{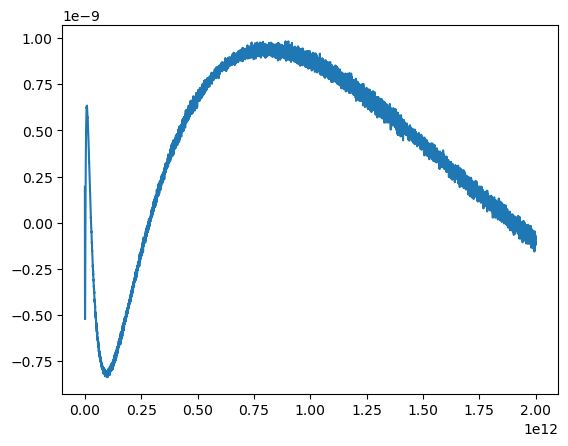}
    
    \end{minipage}
    
    \caption{Final residue in a large domain for 1D Burgers. Upper: weak asymptotics; down: exact asymptotics; left: $\lambda=0.4$; right: $\lambda=0.5$.}
    \label{fig:burgers_final}
\end{figure}

\begin{figure}[htbp]
    \centering
    \begin{minipage}[b]{0.48\textwidth}
        \centering
        \includegraphics[width=\textwidth]{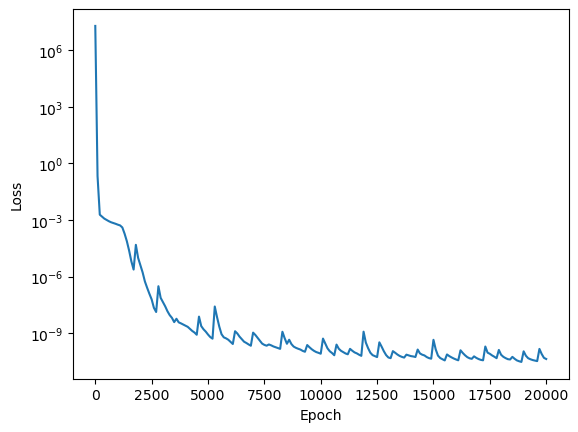}
    
    \end{minipage}
    \hfill
    \begin{minipage}[b]{0.48\textwidth}
        \centering
        \includegraphics[width=\textwidth]{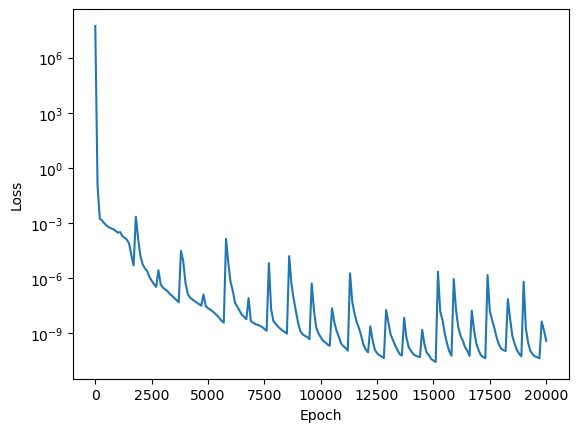}
    \end{minipage}

    \vspace{1em}

    \begin{minipage}[b]{0.48\textwidth}
        \centering
        \includegraphics[width=\textwidth]{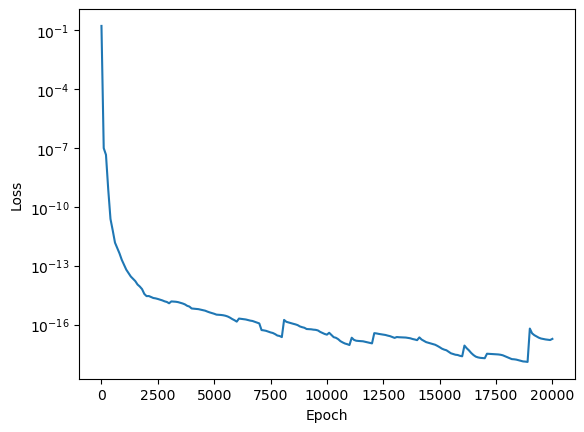}
    \end{minipage}
    \hfill
    \begin{minipage}[b]{0.48\textwidth}
        \centering
        \includegraphics[width=\textwidth]{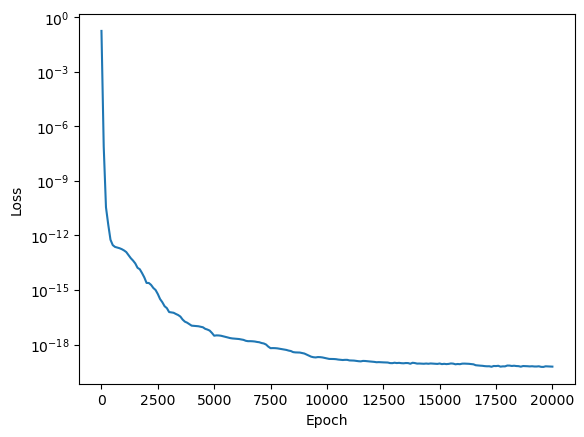}
    
    \end{minipage}
    
    \caption{Trajectory of losses for 1D Burgers. Upper: weak asymptotics; down: exact asymptotics; left: $\lambda=0.4$; right: $\lambda=0.5$.}
    \label{fig:burgers_loss}
\end{figure}

We present the following results of $\lambda=0.4, 0.5$ using weak and exact asymptotics: see Figure \ref{fig:burgers_final} for the equation residue of the solution at the final stage and Figure \ref{fig:burgers_loss} for the evolution of the losses. We are able to achieve high accuracy over a large domain, but using exact asymptotics is preferred for both the smooth and nonsmooth case of $\lambda$.

\subsection{Boussinesq Equation}
For the 2D Boussinesq equation on the half plane, in vorticity form with the self-similar ansatz, we get the following profile equations for $( \Omega,U_1,U_2,\Phi,\Psi)$ as in \cite{wang2023asymptotic}:\begin{align*}
    \Omega+((1+\lambda)(y_1,y_2)^T+(U_1,U_2)^T)\cdot\nabla\Omega&=\Phi,\\
    (2+\partial_{y_1}U_1)\Phi+((1+\lambda)(y_1,y_2)^T+(U_1,U_2)^T)\cdot\nabla\Phi&=-\partial_{y_1}U_2\Psi,\\
    (2+\partial_{y_2}U_2)\Psi+((1+\lambda)(y_1,y_2)^T+(U_1,U_2)^T)\cdot\nabla\Psi&=-\partial_{y_2}U_1\Phi,\\
    \partial_{y_1}U_1+\partial_{y_2}U_2=0, \quad \Omega = \partial_{y_1}U_2-\partial_{y_2}U_1, \quad \partial_{y_1}\Psi &= \partial_{y_2}\Phi,
\end{align*}
where $( \Omega,U_1,\Phi)$ are odd and $(U_2,\Psi)$ are even in $y_1$ and we are in the half plane $y_2\geq0$.

For the boundary conditions, we impose a non-penetration boundary condition $U_2(y_1,0)=0$ along with decaying weak asymptotics at the far field, with Dirichlet boundary conditions $\Phi=\Psi=0$ and Neumann boundary conditions for the velocity field $\nabla(U_1,U_2)^T=0$. For the nondegeneracy condition, we impose $\partial_{y_1}\Omega(0,0)=-1$ and use Taylor expansion to enforce a hard constraint as in Subsection \ref{subsec:hard}. We find that enforcing a hard constraint is much more effective to avoid converging to a trivial solution than enforcing soft constraints.

For each function, we use a $7$-layer MLP with width $30$, hard constraints on parity, and activation function $\mathrm{SiLU}=\frac{x}{1 + e^{-x}}$ to better model the growth at the far field. For sampling, we sample $1000$ points on each boundary of the square $(z_1,z_2)\in[0,30]^2$, and $5000$ points each for the interior and smoothness losses, where we sample  $(z_1,z_2)$ with equal probability uniformly on $[0,30]^2$ and $[0,5]^2$ for the interior loss and with equal probability uniformly on $[0,3]^2$ and $[0,0.5]^2$ for the smoothness loss, ensuring smoothness near the origin. Again, we are computing effectively in a large domain $[0,5\times 10^{12}]^2$ in the $y$ variables.

For optimization, we use Adam for $10000$ epochs with resampling, followed by the optimizer SSBroyden 1 as in \cite{urban2025unveiling} with $40000$ epochs and resampling every 1000 epochs. The learning rate of Adam is set to be $0.001$ for the functions and $0.1$ with $\beta=(0.9,0.9)$ for $\lambda$, with a decay of $0.9$ after $5000$ epochs.

\begin{figure}[htbp]
    \centering
    \begin{minipage}[b]{0.19\textwidth}
        \centering
        \includegraphics[width=\textwidth]{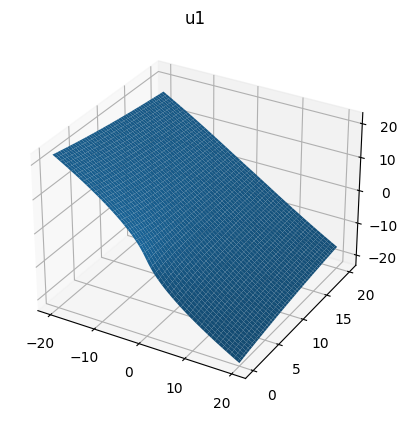}
    
    \end{minipage}
    \hspace{0.1mm}
    \begin{minipage}[b]{0.19\textwidth}
        \centering
        \includegraphics[width=\textwidth]{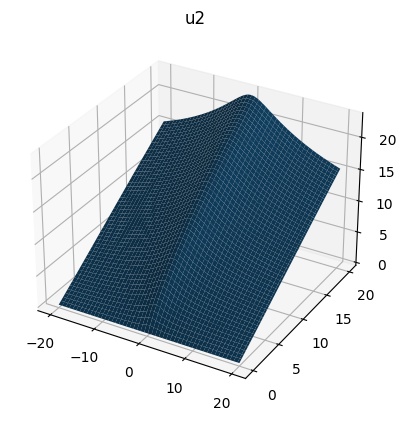}
    \end{minipage}
\hspace{0.1mm}
 \begin{minipage}[b]{0.19\textwidth}
        \centering
        \includegraphics[width=\textwidth]{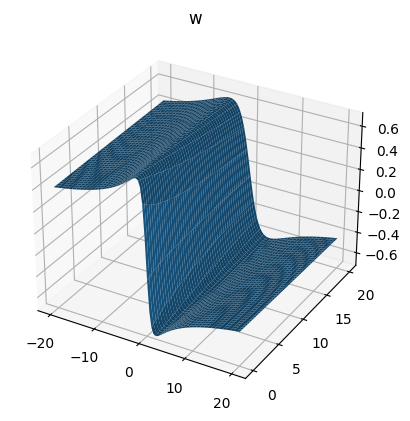}
    
    \end{minipage}
    \hspace{0.1mm}
    \begin{minipage}[b]{0.19\textwidth}
        \centering
        \includegraphics[width=\textwidth]{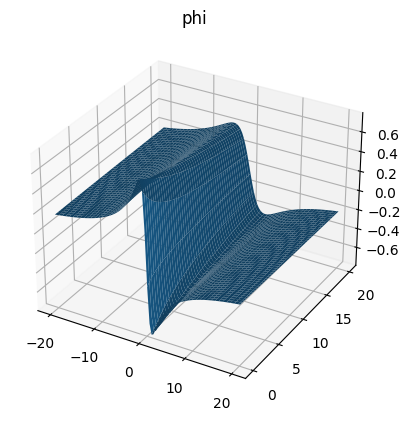}
    
    \end{minipage}
    \hspace{0.1mm}
    \begin{minipage}[b]{0.19\textwidth}
        \centering
        \includegraphics[width=\textwidth]{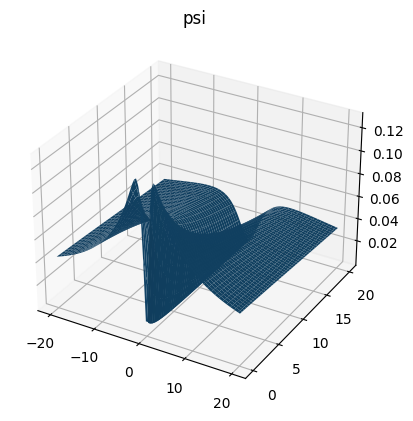}
    \end{minipage}
    
    \caption{Final profiles for 2D Boussinesq}
    \label{fig:bous_prof}
\end{figure}

\begin{figure}[htbp]
    \centering
    \begin{minipage}[b]{0.32\textwidth}
        \centering
        \includegraphics[width=\textwidth]{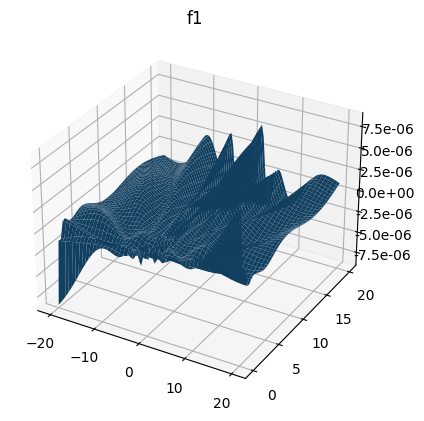}
    
    \end{minipage}
    \hspace{0.1mm}
    \begin{minipage}[b]{0.32\textwidth}
        \centering
        \includegraphics[width=\textwidth]{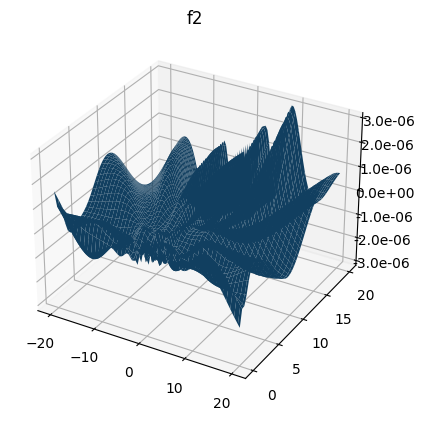}
    \end{minipage}
\hspace{0.1mm}
 \begin{minipage}[b]{0.32\textwidth}
        \centering
        \includegraphics[width=\textwidth]{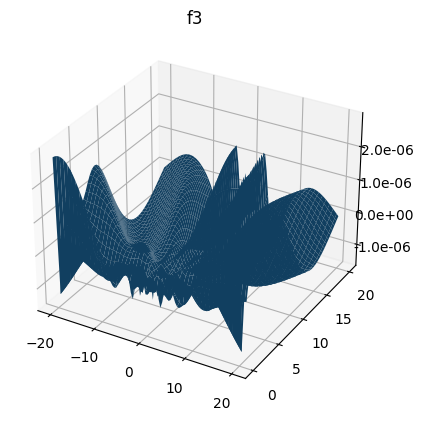}
    
    \end{minipage}
    \vspace{1em}
    \begin{minipage}[b]{0.32\textwidth}
        \centering
        \includegraphics[width=\textwidth]{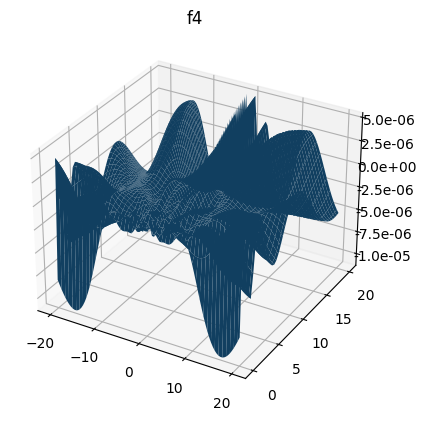}
    
    \end{minipage}
    \hspace{0.1mm}
    \begin{minipage}[b]{0.32\textwidth}
        \centering
        \includegraphics[width=\textwidth]{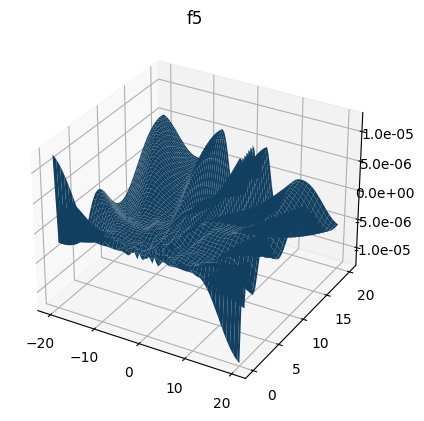}
    
    \end{minipage}
    \hspace{0.1mm}
    \begin{minipage}[b]{0.32\textwidth}
        \centering
        \includegraphics[width=\textwidth]{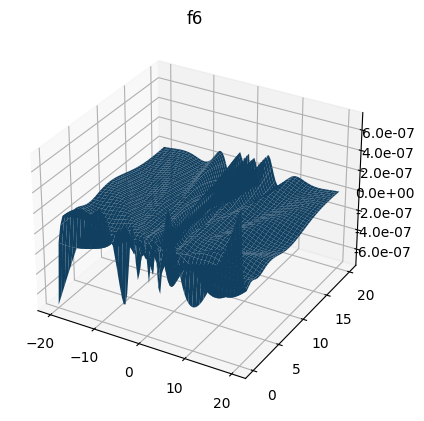}
    \end{minipage}
    
    \caption{Final equation residues for 2D Boussinesq}
    \label{fig:bous_res}
\end{figure}

\begin{figure}
    \centering
\includegraphics[width=0.5\linewidth]{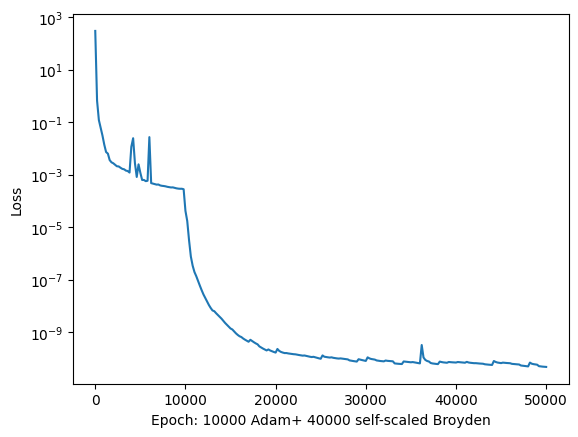}
    \caption{Trajectory of losses for 2D Boussinesq:  
$10000$ Adam iterations followed by $40000$ self-scaled Broyden iterations.}
    \label{fig:traj}
\end{figure}

We present the final profiles and the residue of the PDEs near the origin respectively in Figure \ref{fig:bous_prof} and \ref{fig:bous_res}, and the evolution of losses in Figure \ref{fig:traj}.
Compared to \cite{wang2023asymptotic}, we achieve a training loss of $4$ digits smaller and equation residues of $2$ digits smaller. We remark that due to computational constraints and the cost of a full batch optimizer involving the approximation of the Hessian matrix, this is the largest neural network affordable. We use 10 days of CPU time on a MacPro 2019 with 2.5GHz 28-core Intel Xeon W processor.

    \section{Conclusions and Future Work}
    We demonstrate the importance of enforcing appropriate asymptotics, enforcing hard constraints, and adopting a better optimizer for solving PDEs using neural networks on an infinite domain. We achieve better accuracy for problems crucial to the study of singularity formulations. As a future direction, we believe a better enforcement of far-field asymptotics, formulated as hybrid asymptotics, might have the potential of driving PDE residues to machine precision, potentially amenable to rigorous computer-assisted proofs using the profiles identified by the neural networks. Another direction is to use PINO, the idea of operator learning on a range of scaling parameters, to learn a collection of profiles with different scalings.
    
       \bibliographystyle{plain}
\bibliography{ref}
\end{document}